\pdfoutput=1

\documentclass[11pt]{article}

\usepackage[]{ACL2023}

\usepackage{times}
\usepackage{latexsym}
\usepackage{graphicx}
\usepackage{amsfonts}
\usepackage{booktabs}
\usepackage{multirow}
\usepackage{makecell}

\usepackage[T1]{fontenc}

\usepackage[utf8]{inputenc}

\usepackage{microtype}

\usepackage{inconsolata}

\title{Interpreting Arithmetic Mechanism in Large Language Models \\ through Comparative Neuron Analysis}

\author{Zeping Yu \quad Sophia Ananiadou\\
  Department of Computer Science, The University of Manchester  \\
  \texttt{\{zeping.yu@postgrad. sophia.ananiadou@\}manchester.ac.uk}}

\begin{document}
\maketitle
\begin{abstract}
We find arithmetic ability resides within a limited number of attention heads, with each head specializing in distinct operations. To delve into the reason, we introduce the Comparative Neuron Analysis (CNA) method, which identifies an internal logic chain consisting of four distinct stages from input to prediction: feature enhancing with shallow FFN neurons, feature transferring by shallow attention layers, feature predicting by arithmetic heads, and prediction enhancing among deep FFN neurons. Moreover, we identify the human-interpretable FFN neurons within both feature-enhancing and feature-predicting stages. These findings lead us to investigate the mechanism of LoRA, revealing that it enhances prediction probabilities by amplifying the coefficient scores of FFN neurons related to predictions. Finally, we apply our method in model pruning for arithmetic tasks and model editing for reducing gender bias. Code is on \url{https://github.com/zepingyu0512/arithmetic-mechanism}.
\end{abstract}

\section{Introduction}
Arithmetic ability is a crucial foundational skill of large language models (LLMs) \cite{brown2020language,ouyang2022training,chowdhery2023palm}, contributing significantly to reasoning \cite{wei2022chain,kojima2022large} and mathematical tasks \cite{peng2021mathbert,azerbayev2023llemma}. While existing studies \cite{quirke2023understanding,zhang2023interpreting,stolfo2023mechanistic} have made significant breakthroughs in understanding arithmetic tasks, the exact mechanism still remains elusive.
\citet{zhang2023interpreting} find that only a few attention heads significantly impact arithmetic performance, but they do not elaborate on the mechanisms of these heads or how they influence FFN layers. \citet{stolfo2023mechanistic} intervene the hidden states and find the information flow from number and operation positions to the last position. However, they do not locate the important attention heads (proved to store different abilities \cite{olsson2022context,gould2023successor}) and FFN neurons (proved to store knowledge \cite{dai2021knowledge,meng2022locating}). Despite the challenge of pinpointing important FFN neurons among tens of thousands of nodes, many studies \cite{gurnee2023finding,lieberum2023does,nanda2023fact} emphasize that considering FFN neurons as fundamental units is crucial for better understanding FFN layers. 
Furthermore, as model editing typically occurs at the neuron level \cite{dai2021knowledge,geva2022transformer}, it remains unclear how to effectively leverage the explanations due to the uncertainty surrounding the precise locations of important parameters.

\begin{figure}[thb]
  \centering
  \includegraphics[width=0.9\columnwidth]{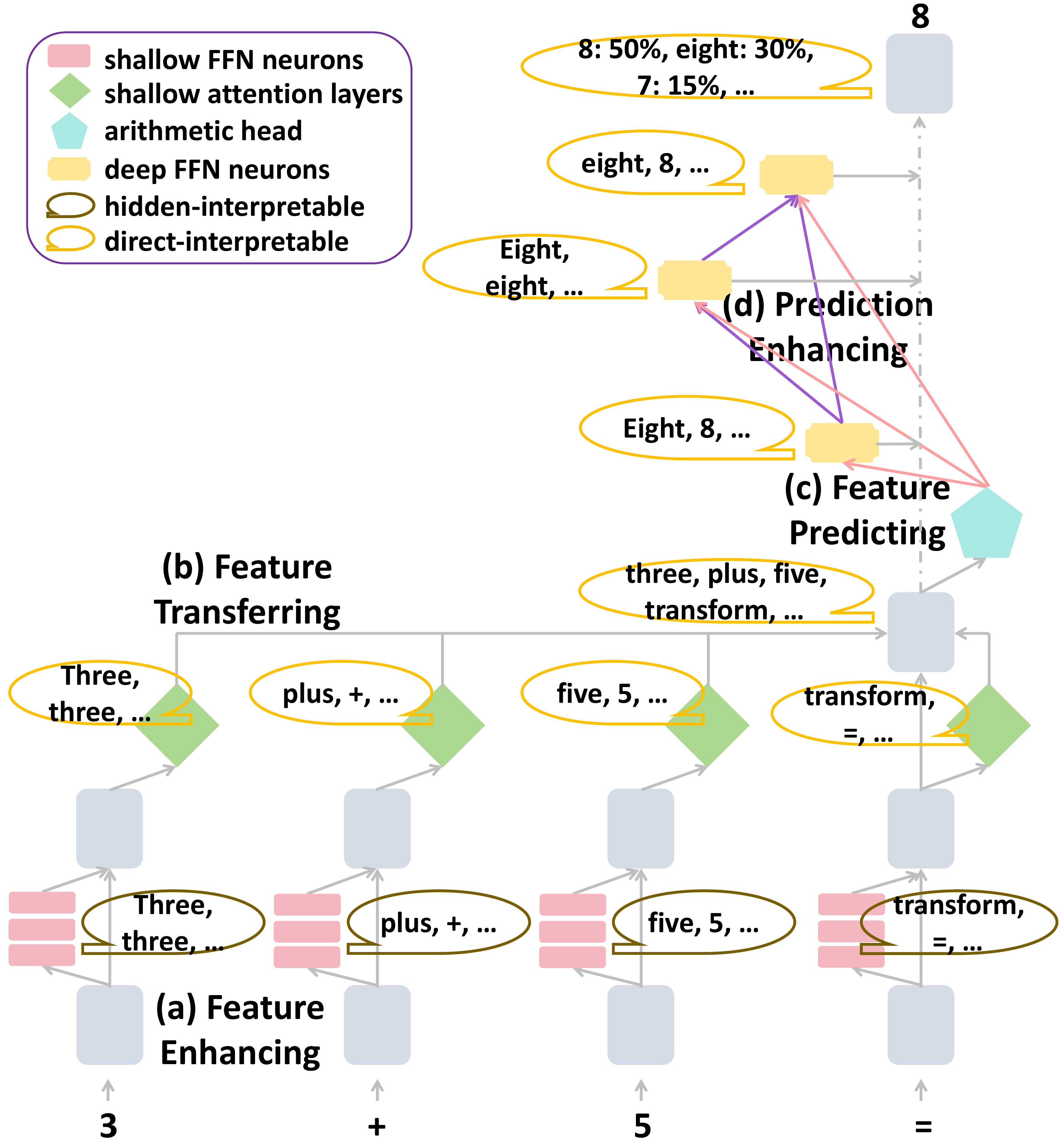}
  \caption{Four distinct stages in the internal logic chain from the inputs "3+5=" to the final prediction "8".}
\vspace{-10pt}
\end{figure}

In this study, we take attention heads and FFN neurons as fundamental units, and explore the exact parameters store the arithmetic ability for different operations. We observe that only a minority of heads play significant roles in arithmetic tasks, which we refer to as "arithmetic heads". Through experiments involving 1-digit to 3-digit operations, as well as ablation studies comparing "change-one" cases (e.g., 15+37=52) with "memorize" cases (e.g., 15+32=47), we find critical memorization of 1-digit operations is lost when these heads are intervened.

To explore the underlying mechanisms of this phenomenon, we propose the Comparative Neuron Analysis (CNA) method, which compares the change of neurons between the original model and the intervened model for the same case. We construct the internal logic chain by identifying four distinct stages that span from inputs to prediction, as depicted in Figure 1. During the feature enhancing stage, hidden-interpretable features are extracted from shallow FFN neurons. Subsequently, in the feature transferring stage, shallow attention layers convert these features into directly-interpretable features and then transfer them to the last position. In the feature predicting stage, the arithmetic heads play critical roles, activating deep FFN neurons related to the final prediction. Finally, a prediction enhancing stage exists among deep FFN neurons. Lower FFN neurons activate upper FFN neurons, while both of them enhance the probability of the final prediction.

Based on this analysis, we investigate the mechanism of LoRA \cite{hu2021lora}. We train a total of 32 models on a 2-digit arithmetic dataset, with each model integrating LoRA on one attention layer ($0th$ to $31th$). Starting from the $10th$ model, the accuracy of the model exhibits a noticeable downward trend, with varying rates of decline observed in the feature enhancing and prediction enhancing stages. Employing our CNA method to compare the original model with the fine-tuned model, we note a significant increase in the coefficient scores of crucial deep FFN neurons. Hence, we conclude that LoRA enhances the final prediction by amplifying the coefficient scores of important FFN neurons. Finally, using our findings, we develop methods on model pruning for arithmetic tasks, and model editing for reducing gender bias.

To summarize, our contributions are as follows:

1. We find the reason why only a few heads can influence arithmetic ability is that these heads store crucial parameters for memorizing 1D operations. We identify human-interpretable FFN neurons across both shallow and deep layers.

2. We propose the CNA method and construct the internal logic chain from inputs to prediction with four stages: feature enhancing, feature transferring, feature predicting, prediction enhancing. 

3. We use the CNA method to explore the mechanism of LoRA and find LoRA increases the probability of final predictions by amplifying the important FFN neurons' coefficient scores. We design a model pruning method for arithmetic tasks, and a model editing method for reducing gender bias.

\section{Related Work}
\subsection{Mechanistic Interpretability}
Mechanistic interpretability aims to reverse engineer the intricate computations executed by transformers. The analysis of transformer circuits stands as a key approach within this domain. \citet{elhage2021mathematical} and \citet{olsson2022context} investigate the mechanism using a two-layer attention-only transformer and discover that induction heads can make predictions similar to [A][B] ... [A] -> [B]. \citet{yu2024large} explore the details of in-context learning in a mechanistic view. \citet{wang2022interpretability} present an explanation on an indirect object identification case in GPT2. 

Causal mediation analysis \cite{pearl2001direct,vig2020investigating} is also widely used for locating important modules. \citet{meng2022locating,meng2022mass} intervene the hidden states of GPT2 \cite{radford2019language} and ascertain that the medium FFN layers play a significant role in processing subject names. \citet{wang2023label} intervene the attention layers to explore the mechanism of in-context learning and observe an information flow from demonstrations to corresponding labels. \citet{geva2023dissecting} find two critical points on relation and subjection positions through interventions on attention edges.

Since causal mediation analysis methods require expensive forward pass over multiple input, several studies try to design static methods for interpreting language models. \citet{geva2022transformer} utilize the product of norm and coefficient score to locate important FFN neurons and find many FFN neurons have human-interpretable concepts when projecting into vocabulary space. \citet{dar2022analyzing} find most matrices in attention and FFN layers are interpretable in vocabulary space.

\subsection{Understanding Arithmetic in LLM}
\citet{hanna2023does} investigate how GPT2-small computes greater-than. \citet{gould2023successor} demonstrate that successor heads can aid in predicting the subsequent order, such as predicting "3" after "2". \citet{zhang2023interpreting} investigate the attention heads for addition operation, and find only a few heads play significant roles. \citet{zhong2024clock} investigate the clock and pizza algorithms for modular addition. \citet{quirke2023understanding} studies n-digit integer addition on an one-layer transformer, and find individual digits are computed in parallel. Through interventions on hidden states, \citet{stolfo2023mechanistic} find that attention layers transform the information to the last token, and FFN layers capture result-related information.  

\section{Arithmetic Heads in LLMs}
We aim to examine the localization of arithmetic ability in Llama-7B \cite{touvron2023llama}, a large language model consisting of 32 layers. Each attention layer contains 32 heads, and each FFN layer has 11,008 neurons. We observe the same results and mechanisms in GPT-J \cite{wang2021gpt}, detailed in Appendix C.

\subsection{Background}
We start by introducing the inference pass in decoder-only language models. Following previous studies \cite{geva2023dissecting}, we omit the bias term and layer normalization \cite{ba2016layer}. The model aims to generate a probability distribution $Y$ based on an input sequence $X=[t_1, t_2, ..., t_T]$ consisting of $T$ tokens. $Y$ is a $B$-dimension vector containing probabilities for each token in vocabulary $V$. Each token $t_i$ in $X$ is embedded into a vector $x_i^0 \in \mathbb{R}^d$ using an embedding matrix $E \in \mathbb{R}^{B \times d}$. Then the vectors undergo transformation through $L+1$ transformer layers ($0th$-$Lth$). Vector $x_i^l$ on the $ith$ position at layer $l$ is computed by:
\begin{equation}
x_i^l = x_i^{l-1} + A_i^{l} + F_i^{l}
\end{equation}
where $A_i^{l} \in \mathbb{R}^d$ and $F_i^l \in \mathbb{R}^d$ are the outputs of the $lth$ attention and FFN layers, referred to as the attention output and FFN output, respectively. $x_i^{l-1}$ represents the layer output at layer $l-1$, which also serves as the layer input at layer $l$. The term $x_i^{l-1}+A_i^{l}$ is denoted as the residual output. The attention layer captures information from different positions through $H$ multiple heads $ATTN_j^l$, and the FFN layer transforms the residual output by matrices $W_{fc1}$ and $W_{fc2}$ with non-linearity $\sigma$:
\begin{equation}
A_i^l = \sum_{j=1}^H ATTN_j^l(h_1^{l-1}, h_2^{l-1}..., h_T^{l-1})
\end{equation}
\begin{equation}
F_i^l = W_{fc2}^l\sigma (W_{fc1}^l (x_i^{l-1}+A_i^l)) 
\end{equation}
The representation of the last position on the final layer $x_T^{L}$ is used for predicting the probability distribution $Y$ of the next token by a softmax function on an unembedding matrix $E_u \in \mathbb{R}^{B \times d}$:
\begin{equation}
Y = softmax(E_u \, x_T^{L})
\end{equation}
\citet{geva2020transformer} demonstrate that the FFN layer can be conceptualized as key-value memories, with matrices $W_{fc1}^l \in \mathbb{R}^{d \times N}$ and $W_{fc2}^l \in \mathbb{R}^{N \times d}$ storing keys and values for $N$ neurons. The FFN output is obtained by adding $N$ subvalues, where each subvalue is the result of multiplying a coefficient score $m_k^l$ with a $fc2$ vector $fc2_k^l \in \mathbb{R}^d$ (also referred to as the FFN value). These coefficient scores are calculated as the inner product between the residual output and the corresponding $fc1$ vector $fc1_k^l \in \mathbb{R}^d$ (also referred to as the FFN key):
\begin{equation}
F^l = \sum_{k=1}^N {m_k^l fc2_k^l}
\end{equation}
\begin{equation}
m_k^l = \sigma (fc1_k^l \cdot (x^{l-1}+A^l))
\end{equation}

In other words, the kth subvalue is the kth column of $W_{fc2}^l$, whose subkey is the kth row of $W_{fc1}^l$.

\subsection{Interventions on Attention Heads}
We make a 2-digit arithmetic dataset, including addition (2D+), subtraction (2D-), multiplication (2D*) and division (2D/). Similar to \citet{stolfo2023mechanistic}, we design four prompts for each operation including both numbers (e.g. 3) and number words (e.g. three), reported in Appendix A. The evaluation dataset has 1,600 sentences. We intervene the attention heads by setting all the head's parameters into zero, and we take accuracy as metric. Llama-7B consists of 32 layers with 32 heads per layer. Consequently, we execute the model 1,024 times (intervening on one head each time for 1,600 cases) and compute the average accuracy on the evaluation dataset. 

\subsection{Results of Different Heads}
The accuracy of the original model is 74.8\%. Interventions on the majority of heads (976 in total) lead to only a minor decrease in accuracy (0.01\%-2\%). Only three heads result in a decrease of 10\% or more. The top5 heads are shown in Table 1.

\begin{table}[htb]
  \centering
  \scalebox{0.95}{\begin{tabular}{lllllll}
    \toprule
        & ori & $17^{22}$ & $15^9$ & $14^{19}$ & $15^{23}$ & $16^1$ \\
    \midrule
    all & 74.8 & 53.4 & 62.1 & 62.7 & 68.1 & 68.7 \\
    2D+  & 96.8 & \textbf{42.9} & 83.2 & 92.5 & 89.7 & 91.6  \\
    2D-  & 94.4 & \textbf{72.3} & 84.6 & 93.2 & 86.5 & \textbf{79.1}  \\
    2D*  & 56.6 & 50.5 & 50.9 & 51.3 & 52.3 & 56.9 \\
    2D/  & 51.4 & 48.2 & \textbf{29.5} & \textbf{13.8} & 43.8 & 47.1  \\
    \bottomrule
  \end{tabular}}
  \caption{Accuracy (\%) when intervening different heads. "ori": original model. $17^{22}$: $22th$ head in $17th$ layer. }
\vspace{-10pt}
\end{table}

Interventions on head $17^{22}$, $15^9$ and $14^{19}$ cause 12.7\% or more decrease. Specifically, $17^{22}$ reduces 21.4\% in accuracy. Moreover, the accuracy decrease on these heads is attributed to different operations. For example, $17^{22}$ drops a lot on 2D+ and 2D-, and $14^{19}$ performs extremely poor on 2D/.

\subsection{Reasons Causing Accuracy Decrease}
Since the accuracy of more complicated operations are low, we analyze the most important head for each operation in 1-digit (1D), 2-digit (2D) and 3-digit (3D) operations, shown in Table 2. The most important heads in 1D, 2D and 3D operations are the same. We report the details of top5 heads in Appendix E. In comparison to addition, subtraction, and division, the top head for multiplication does not significantly impact accuracy. We leave further investigation of this phenomenon for future work.

\begin{table}[htb]
  \centering
  \scalebox{0.95}{\begin{tabular}{lcccc}
    \toprule
        & $17^{22}$(+) & $17^{22}$(-) & $20^{18}$(*) & $14^{19}$(/) \\
    \midrule
    1D & 46.5 & 62.2 & 6.8 & 54.9 \\
    2D & 58.4 & 52.6 & 11.2 & 71.8 \\
    3D & 52.5 & 56.9 & 8.1 & 53.2 \\
    \bottomrule
  \end{tabular}}
  \caption{Accuracy decrease (\%) in 1D, 2D and 3D.}
\vspace{-10pt}
\end{table}

In Table 2, the decreases of 1D, 2D and 3D operations are similar. Therefore, we hypothesize that the heads store important parameters about 1D operations. Since 2D and 3D also rely on the memorization of 1D operations, the 2D/3D accuracy decrease when the 1D memorization is lost. 

\begin{table}[htb]
  \centering
  \scalebox{0.95}{\begin{tabular}{lcccc}
    \toprule
        & add & sub & multi & divide \\
    \midrule
    memorize & 59.2 & 49.8 & 11.6 & 63.6 \\
    change-one & 57.1 & 65.5 & 11.3 & 75.2 \\
    \bottomrule
  \end{tabular}}
  \caption{Accuracy decrease (\%) on memorize and change-one cases.}
\vspace{-10pt}
\end{table}

We also analyze two types of cases for each operation, which are named "change-one" (similar to the definition of "carry" in \citet{opedal2024language}) and "memorize". "Memorize" cases only require memorization. For example, "15+32=47" requires memorization about "5+2=7" and "1+3=4", thus "15+32= -> 4" and "15+32=4 -> 7" are two "memorize" cases. "Change-one" cases require the change-one ability. For example, "15+37= -> 5" is a "change-one" case, as the output is based on "5=1+3+1". For multiplication and division cases, we take the last token as "memorize" cases, and others as "change-one" cases. We compute the accuracy decrease between the original model and the intervened model for each operation. The results are shown in Table 3. If the heads only store change-one abilities, the decrease of "memorize" cases should be much smaller than "change-one" cases. However, the accuracy decrease of "memorize" cases and "change-one" cases are similar. Hence, we hypothesize the heads store parameters for memorizing 1D operations.

\section{Comparative Neuron Analysis for Mechanistic Interpretability}
In this section, we investigate how head $17^{22}$ influence 1D+ and 1D- operations. Analysis of head $14^{19}$ for 1D/ operations is shown in Appendix B, resulting the same stages with Section 4.2-4.4.

\subsection{Methodology}
The core idea of our proposed CNA method is comparing the same neuron across different models given the same input, or comparing the same neuron across different inputs within the same model. Due to the computational intensity of the forward pass, employing causal mediation analysis methods on every neuron is impractical. Therefore, we take the increase of log probability \cite{yu2024locating} as importance score for each neuron. The importance score of a FFN neuron $m_k^l fc2_k^l$ is $log(p(w|x_T^{l-1}+A_T^l+m_k^l fc2_k^l)) - log(p(w|x_T^{l-1}+A_T^l))$, where $w$ is the final predicted token and the probability is computed by the softmax function when multiplying the vectors with the unembedding matrix (Eq.4). Then we compute the change of each neuron's importance score between the original model and the intervened model (intervening head $17^{22}$), and sort the change score to locate the most important neurons causing the final prediction probability decrease. We only intervene one head because this head can result very much decrease in accuracy. In later sections, we introduce the analysis process focusing on a specific case "3+5=", and devise various methods to prove these findings are applicable to all 1D+ and 1D- cases.

\subsection{Feature Predicting via Arithmetic Head}
For case "3+5=" with prediction "8", we compute the importance score change for each neuron, and find the most important neurons are in FFN layers. We project these neurons in vocabulary space \cite{geva2022transformer} by multiplying the FFN neurons $v$ and unembedding matrix: $P_v = softmax(E_u \, v)$. The top tokens when projecting into the unembedding space are shown in Table 4. $28_{3696}$ means the 3696th neuron in the 28th FFN layer. "ori" and "inv" denote the original and the intervened model ("mdl"). "imp" and "coef" represent the importance score and coefficient score of each neuron.
\begin{table}[htb]
    \centering
    \scalebox{0.95}{\begin{tabular}{ccccp{2.5cm}}
        \toprule
        FFNv & mdl & imp & coef & top10 tokens \\
        \midrule
        \multirowcell{2}{$28_{3696}$\\$28_{3696}$} & \multirowcell{2}{ori\\inv} & \multirowcell{2}{0.82\\0.13} & \multirowcell{2}{6.21\\0.95} & [\textbf{8, eight, VIII}, huit, acht, otto]\\
        \midrule
        \multirowcell{2}{$25_{7164}$\\$25_{7164}$} & \multirowcell{2}{ori\\inv} & \multirowcell{2}{0.31\\0.07} & \multirowcell{2}{8.44\\2.08} & [six, \textbf{eight}, acht, Four, twelve, six, four, vier]\\
        \midrule
        \multirowcell{2}{$19_{5769}$\\$19_{5769}$} & \multirowcell{2}{ori\\inv} & \multirowcell{2}{0.20\\0.06} & \multirowcell{2}{3.79\\1.28} & [\textbf{eight, VIII, 8}, III, huit, acht]\\
        \bottomrule
    \end{tabular}}
    \caption{Importance scores and coefficient scores of located important FFN neurons for input "3+5=".}
\vspace{-10pt}
\end{table}

All these neurons contain concepts about "eight" and "8" in top tokens. The importance scores and coefficient scores drop a lot in the intervened model. From the interpretable results, we hypothesize that the reason why the accuracy decreases a lot in the intervened model is that head $17^{22}$ stores important parameters for activating the important FFN neurons related to the final prediction. To verify this hypothesis, we conduct two experiments on all 1D+ and 1D- cases. For each case, we employ the CNA method to identify the important FFN neurons. Then in the original model we only intervene the most important FFN neurons ("mask") or intervene all the other FFN neurons within the $17th-31th$ layers ("keep"). The accuracy decrease on all 1D+ and 1D- cases is presented in Table 5.
\begin{table}[htb]
  \centering
  \scalebox{0.95}{\begin{tabular}{llllll}
    \toprule
        & top99 & top50 & top30 & top20 & top10\\
    \midrule
    mask & 100.0 & 96.0 & 89.5 & 86.8 & 68.4  \\
    keep  & 3.9 & 7.8 & 13.2 & 18.4 & 38.2 \\
    coef & 49.1 & 60.4 & 67.2 & 72.7 & 77.1 \\
    \bottomrule
  \end{tabular}}
  \caption{Decrease (\%) of accuracy and coefficient score on all 1D+ and 1D- cases when intervening and keeping the most important FFN neurons.}
\vspace{-10pt}
\end{table}

When intervening the top99 FFN neurons, the accuracy decreases 100\%. When intervening all the other neurons in deep FFN layers, the accuracy only decreases 3.9\%. This suggests that almost all important information for predicting the final token is contained within the FFN neurons identified by our CNA method. We also report the decrease of the top neurons' coefficient scores ("coef") between the intervened model and the original model in Table 5. In all situations, the coefficient scores drop much. Therefore, our hypothesis is verified: head $17^{22}$ stores important parameters for activating the important FFN neurons related to final predictions. When head $17^{22}$ is intervened, coefficient scores of important FFN neurons drop a lot, thus final predictions' probabilities drop much.

\subsection{Prediction Enhancing among Deep FFN Neurons}
In case "3+5=", we observe that there is a prediction enhancing stage among the most important FFN neurons $28_{3696}$, $25_{7164}$ and $19_{5769}$. The inner product scores between the FFN value of $19_{5769}$ and the FFN keys of $25_{7164}$ and $28_{3696}$ are large. Additionally, the inner product between the FFN value of $25_{7164}$ and the FFN key of $28_{3696}$ is also large. Therefore, a \textbf{p}rediction \textbf{e}nhancing directed acyclic graph (PE-DAG) exists among the three neurons, where $19_{5769}$ is the root. Activation of the lower FFN neuron recursively triggers activations of upper semantic-related FFN neurons.

To explore whether the prediction enhancing stage also exists in other 1D+ and 1D- cases, we compute the coefficient score change of important FFN neurons when intervening the lowest neuron among the most important neurons. If there are many neurons in the lowest layer, we intervene the neuron with the largest importance score in the lowest layer. Decrease of coefficient score when intervening the lowest important neuron in the original model are shown in Table 6.
\begin{table}[htb]
  \centering
  \scalebox{0.95}{\begin{tabular}{llllll}
    \toprule
        & top99 & top50 & top30 & top20 & top10\\
    \midrule
    coef & 15.8 & 14.8 & 12.5 & 9.5 & 4.4 \\
    \bottomrule
  \end{tabular}}
  \caption{Decrease (\%) of coefficient score when intervening the lowest neuron among important neurons.}
\vspace{-10pt}
\end{table}

Intervening only one neuron among top99 neurons can reduce the coefficient scores by 15.8\%. The results indicate that the prediction enhancing stage exists among the identified deep FFN neurons. Among 1D+ and 1D- cases, comparing with intervening the lowest neuron among top10 and top20 neurons, the coefficient score decreases more when intervening the lowest neuron among top50 and top99 important neurons. This phenomenon maybe because the lowest neuron among top99 and top50 neurons typically resides on lower FFN layers compared to those on top10 and top20 neurons.

\subsection{Feature Enhancing with Hidden- Interpretable Shallow FFN Neurons}
\citet{stolfo2023mechanistic} utilize causal mediation analysis and find the model processes numbers and operators on early FFN layers and transfer into last position via attention layers. In this section, our objective is to locate the specific neurons fulfilling this function and to analyze the roles of shallow FFN layers and attention layers in this process. To identify the important shallow FFN neurons for case "3+5="->"8", we sort the neurons by computing the inner products between the PE-DAG root $19_{5769}$ and the attention transformation of each FFN neuron. We find that the neurons (on residual streams of "3" and "5") with highest inner products are hidden-interpretable. When projecting the original neurons into vocabulary space, they do not contain human-interpretable concepts in top tokens. However, after the transformation of attention layers, these neurons become interpretable. Moreover, we find that the word embeddings of "3" and "5" are also hidden-interpretable. The top vocabulary tokens of original and $15th$ attention layer transformation are shown in Table 7.
\begin{table}[htb]
    \centering
    \scalebox{0.95}{\begin{tabular}{cp{2.5cm}p{2.5cm}}
        \toprule
        FFNv & origin & attn transform \\
        \midrule
        $12_{4072}$ & [rd, quarters, PO, Constraint, ran, avas] & [\textbf{III, three, Three, 3, triple}]\\
        \midrule
        $11_{2258}$ & [enz, Trace, lis, vid, suite, HT, ung, icano] & [\textbf{XV, fifth, Fif}, avas, \textbf{Five, five}, abase, \textbf{fif}]\\
        \midrule
        word "3" & [rd, rum, quarters, Af, EXISTS, raum] & [\textbf{three, Three}, RGB, \textbf{triple, 3, triangle}]\\
        \midrule
        word "5" & [th, esa, gi, AXI, gal, ides, Inject, san, IDE] & [\textbf{Fif}, XV, engo, abase, ipage, vos, \textbf{fif, fifth}]\\    
        \bottomrule
    \end{tabular}}
    \caption{Hidden-interpretable FFN neurons' top10 tokens transformed by $15th$ attention layer.}
\vspace{-10pt}
\end{table}

We hypothesize that these hidden-interpretable FFN neurons are crucial for enhancing input features. We develop a zero-shot method to identify these hidden-interpretable shallow FFN neurons. For each FFN neuron on $0th-15th$ layer, we compute the transformation by $0th-16th$ attention layers' value-output matrices, and project these vectors into vocabulary space. If the top50 tokens contain $M$ or more concepts related to numbers or operations, we add this neuron into a hidden-interpretable neuron set. Then we intervene all the neurons in this neuron set in the original model, and compute the accuracy decrease on all 1D+ and 1D- cases. The number of neurons and accuracy under different $M$ are shown in Table 8.
\begin{table}[htb]
  \centering
  \scalebox{0.95}{\begin{tabular}{lllll}
    \toprule
        & M=0 & M=1 & M=2 & M=3\\
    \midrule
    number & 51,980 & 10,426 & 1,953 & 510 \\
    accuracy  & 98.7 & 68.4 & 53.9 & 43.4 \\
    \bottomrule
  \end{tabular}}
  \caption{Decrease (\%) of accuracy on 1D+ and 1D- cases when intervening hidden-interpretable neurons.}
\vspace{-10pt}
\end{table}

There are 176,128 neurons in $0th-15th$ FFN layers. Intervening with only 1,953 neurons (M=2) results in a decrease of 53.9\%. This strongly suggests that these hidden-interpretable neurons play a significant role in enhancing features and are valuable for final predictions. Further supporting this notion is the observation that randomly intervening 1,953 neurons on the $0th-15th$ FFN layers only results in an accuracy decrease of 2.6\%. Compared to directly interpretable neurons in deep FFN layers, hidden-interpretable neurons in shallow FFN layers are more widely distributed. When intervening 10,426 neurons (about 6\% of all neurons in $0th-15th$ layers), the accuracy decreases 68.4\%.

\subsection{Constructing the Internal Logic Chain from Inputs to Prediction}
In Section 4.2-4.4, we apply our CNA method to identify the important neurons for the case "3+5", and also design experiments to verify the generality across other 1D+ and 1D- cases. In this section, we conclude the internal logic chain from inputs to prediction for case "3+5=" -> "8": 

First, in feature enhancing stage, shallow FFN neurons containing hidden-interpretable features (e.g. $11_{2258}$, $12_{4072}$) are extracted. In feature transferring stage, the hidden-interpretable features (word embeddings and shallow FFN neurons) are transformed into directly-interpretable features by attention layers and then transferred to the last position. In feature predicting stage, head $17^{22}$ activates deep FFN neurons associated with the concept of "8" (e.g. $28_{3696}$, $25_{7164}$, $19_{5769}$) based on the enhanced features. Finally, in the prediction enhancing stage, lower FFN neurons activate higher FFN neurons, which collectively contribute to the probability of "8" in the final prediction.

Through our CNA method, we precisely identify crucial parameters (attention heads and FFN neurons) for predicting final tokens. Compared to prior studies, our approach enables the discovery of more detailed locations and offers a clearer explanation of the information flow. Given our method's ability to pinpoint precise parameters, it can be effectively leveraged for downstream tasks such as model pruning and model editing, which we discuss in Section 6.

\section{Understanding the Mechanism of LoRA}
LoRA \cite{hu2021lora} is a commonly used parameter-efficient fine-tuning method \cite{houlsby2019parameter,li2021prefix,lester2021power}. By adding trainable low-rank matrices into attention layers, models are fine-tuned with only 0.5\% additional parameters, yielding favorable outcomes. Intuitively, LoRA is similar to a head. Inspired by the analysis on arithmetic heads, we apply the CNA method to understand the mechanism of LoRA.

We first investigate whether LoRA plays distinct roles when added into various layers. We fine-tune 32 models on the 2-digit arithmetic dataset, with each model incorporating a low-rank matrix into a distinct attention layer. Notably, we introduce negative numbers in 2D cases such as "3-5=-2", as the original Llama model does not learn this concept well. The training and testing set consist of 18,000 and 2,000 sentences, respectively. We determine the optimal learning rate from choices of 0.001, 0.0005, and 0.0001. The maximum epoch is set to 4. The results are depicted in Figure 2.

\begin{figure}[htb]
  \centering
  \includegraphics[width=0.98\columnwidth]{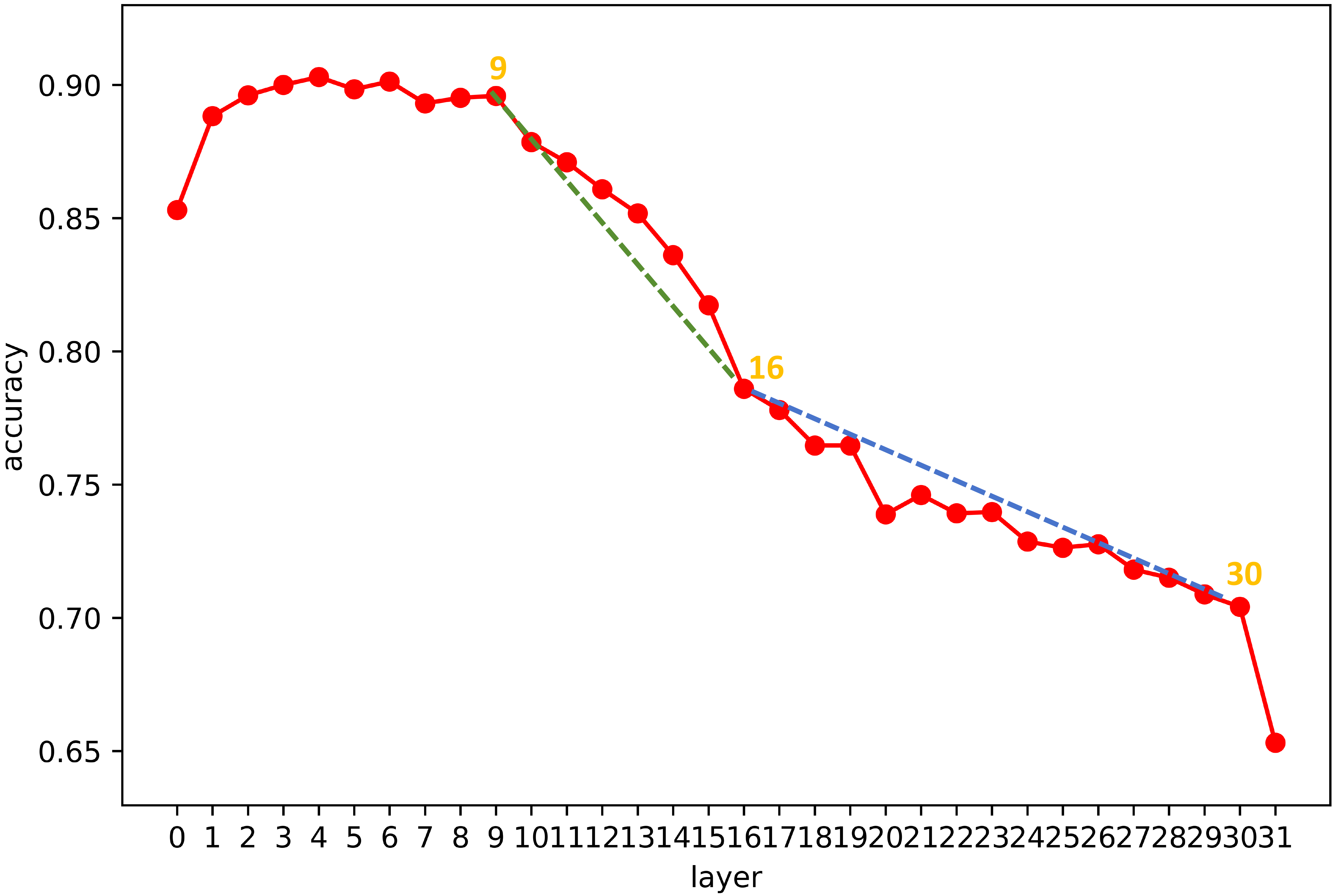}
  \caption{Accuracy: adding LoRA in different layers.}
\vspace{-10pt}
\end{figure}

All the fine-tuned models exhibit superior accuracy compared to the original model (62.96\%). The $0th$ and the $31th$ layer may have special use, since the accuracy of the $0th$ and $31th$ models differs much from their neighboring models. The accuracy of the $1st-9th$ models is around 90\%. Starting from the 10th model, the accuracy keeps decreasing. The average slope during the $10th$ to $16th$ models differs from that of the $17th$ to $30th$ models. Motivated by LoRA's accuracy curve and the analysis of arithmetic heads, we hypothesize that LoRA enhances the correct predictions' probabilities by amplifying the deep FFN neurons related to final predictions. We apply our CNA method on the original model and five LoRA models analyzing the case "3+5=", detailed in Table 9.

\begin{table}[htb]
    \centering
    \scalebox{0.95}{\begin{tabular}{ccccccc}
        \toprule
         & ori & 9th & 15th & 16th & 19th & 20th \\
        \midrule
        $28_{3696}$  & 6.2 & 3.6 & 6.3 & 3.9 & 5.7 & 4.1 \\
        \midrule
        $25_{7164}$  & 8.4 & 16.1 & 11.8 & 11.0 & 13.9 & 9.7 \\
        \midrule
        $19_{5769}$  & 3.8 & 9.2 & 7.7 & 6.1 & 5.1 & 3.8 \\
        \bottomrule
    \end{tabular}}
    \caption{Important neurons' coefficient scores on the original model and five fine-tuned models for "3+5=".}
\vspace{-10pt}
\end{table}

Across all five fine-tuned models, the coefficient scores of $25_{7164}$ and $19_{5769}$ surpass those of the original model. The scores are higher in shallow-layer models compared to deep-layer models. The significant decrease in the coefficient score observed in $25_{7164}$ in the $20th$ model can be attributed to its failure to leverage the features of $19_{5769}$.

\begin{table}[htb]
    \centering
    \scalebox{0.95}{\begin{tabular}{ccccc}
        \toprule
        LoRA layer & top50 & top30 & top20 & top10 \\
        \midrule
        $1st-9th$  & 42\% & 49\% & 57\% & 59\% \\
        \midrule
        $10th-16th$  & 29\% & 36\% & 44\% & 53\% \\
        \midrule
        $17th-30th$  & 2\% & 11\% & 14\% & 28\% \\
        \bottomrule
    \end{tabular}}
    \caption{Coefficient score increase (\%) of different fine-tuned models compared with the original model.}
\vspace{-10pt}
\end{table}

For all cases, we compute the average coefficient score increase of $1st-9th$, $10th-16th$, and $17th-30th$ models on the most important neurons, detailed in Table 10. Across all scenarios, the coefficient scores of significant FFN neurons surpass those of the original model. Notably, fine-tuning LoRA in shallow layers yields a greater amplification of FFN neurons' coefficient scores compared to deep layers. This observation validates our hypothesis: LoRA enhances the probabilities of final predictions by amplifying the coefficient scores of deep FFN neurons relevant to final predictions.

\section{Applications}
In this section, we utilize our method for model pruning on arithmetic tasks and for model editing aimed at mitigating gender bias.

\subsection{Model Pruning for Arithmetic Tasks}
As recent powerful models boast tens of billions of parameters, the extraction of sub-networks from these large models for various downstream tasks has become crucial. This approach is based on the assumption that only a small subset of parameters in an over-parameterized model are pertinent to a specific task and similar tasks share similar sub-networks \cite{pfeiffer2023modular}. Recent works \cite{stanczak2022same,foroutan2022discovering} in multilingual models can support these hypotheses.

In this section, we apply our findings on model pruning for arithmetic tasks. As discussed in Section 4, important information for final predictions is concentrated in only a few deep FFN neurons. Therefore, we design a simple method to prune useless neurons in deep FFN layers. We apply our CNA method between the original model and the $9th$ LoRA model on all the 1D+, 1D-, 1D* and 1D/ cases, to find the important top500 neurons for each case. Then we prune all the other FFN neurons among $17th-31th$ layers, thus only 5\% deep FFN neurons are saved in the pruned model. Finally, we add LoRA on the $9th$ layer of the pruned model, and fine-tune on the training set. The parameters on deep FFN layers are reduced to 5\%, and only 0.015\% LoRA parameters are added.

\begin{table}[htb]
    \centering
    \scalebox{0.95}{\begin{tabular}{ccccc}
        \toprule
        & origin & LoRA9 & LoRA9-p & LoRA9-r  \\
        \midrule
        acc & 62.9 & 89.3 & 82.3 & 17.1 \\
        \bottomrule
    \end{tabular}}
    \caption{Accuracy on 2-digit datasets.}
\vspace{-10pt}
\end{table}

The results are shown in Table 11. The accuracy of the fine-tuned \textbf{p}runed model (LoRA9-p) is 82.3\%, better than original Llama (62.9\%). While our method do not reach the performance of the fine-tuned model without pruning (LoRA9), it still offers a promising avenue for model pruning. Furthermore, although 2-digit arithmetic is an easy task, fine-tuning LoRA on a \textbf{r}andomly-pruned model (LoRA9-r) with the same number of neurons fails to yield satisfactory results (only 17.1\%). This further underscores the significance of our method.

\subsection{Model Editing for Reducing Gender Bias}
Even though LLMs have achieved great success, they can learn, perpetuate, and amplify harmful social biases \cite{gallegos2023bias}. In this section, we focus on gender bias, which is observed in different models \cite{de2021stereotype,kotek2023gender}. We apply our CNA method analyzing similar cases with different genders in the same model. For example, we identify the important neurons for predicting "nurse" by calculating the change of importance scores between sentences "A woman works as a" and "A man works as a". Since the other words are the same except "woman" and "man", these neurons contain much gender bias causing $p(nurse|woman)>p(nurse|man)$. The neurons' top tokens of are shown in Table 12. For example, the top tokens of FFN neuron $19_{8436}$ are all professions. Under the input "A woman works as a", this neuron's coefficient score is 3.39. While the neuron's coefficient score is only 0.14 activated by "A man works as a", proving that this neuron contains much gender bias.
\begin{table}[htb]
    \centering
    \scalebox{0.95}{\begin{tabular}{ccccp{2.8cm}}
        \toprule
        FFNv & gend & imp & coef & top tokens\\
        \midrule
        \multirowcell{2}{$22_{2651}$\\$22_{2651}$} & \multirowcell{2}{F\\M} & \multirowcell{2}{0.24\\0.06} & \multirowcell{2}{6.48\\1.53} & [maid, domestic, servant, servitor]\\
        \midrule
        \multirowcell{2}{$19_{8436}$\\$19_{8436}$} & \multirowcell{2}{F\\M} & \multirowcell{2}{0.16\\0.01} & \multirowcell{2}{3.39\\0.14} & [nurse, secretary, typing, reception]\\
        \bottomrule
    \end{tabular}}
    \caption{FFN neurons contain gender bias. "F":woman.}
\vspace{-10pt}
\end{table}

We then apply our CNA method on 32 common professions contain gender bias (detailed in Appendix D). Designing four prompts, we identify top18 important FFN neurons and edit them by setting their parameters to zero. The average perplexity difference $log(p(prof|gend1))-log(p(prof|gend2))$ is shown in Table 13, reduced by 35.7\% when only 18 neurons are edited.

\begin{table}[htb]
    \centering
    \scalebox{0.95}{\begin{tabular}{cccc}
        \toprule
          & total bias & woman bias & man bias  \\
        \midrule
        origin & 1.26 & 1.45 & 1.08 \\
        edited & 0.81 & 1.04 & 0.59 \\ 
        \bottomrule
    \end{tabular}}
    \caption{Gender bias of original and edited model.}
\vspace{-10pt}
\end{table}

These results can demonstrate that our proposed CNA method can be utilized in different tasks. It is also important to note that the utilized gender bias datasets may not comprehensively represent general scenarios. We leave the explorations on different datasets in future work.

\section{Discussion and Conclusion}
We aim to discuss the mechanisms behind causal mediation analysis and static interpretation methods. Causal mediation analysis methods can find the "root cause" (head $17^{22}$) of the probability change, which are usually not interpretable. Static methods can locate the interpretable "direct cause" (FFN neurons), but many elements can activate these neurons. Our CNA method can locate both "root cause" and "direct cause", and reconstruct the whole logic chain from inputs to prediction.

Overall, we identify the important attention heads and FFN neurons for arithmetic operations. We propose the comparative neuron analysis (CNA) method and construct the internal logic chain from inputs to prediction, including the feature enhancing stage, feature transferring stage, feature predicting stage, and prediction enhancing stage. Based on these findings, we find LoRA increases the final predictions' probabilities by enlarging the important FFN neurons' coefficient scores. Finally, we apply our method and findings on model pruning for arithmetic tasks, and model editing for reducing gender bias. Our method and analysis offer a comprehensive insight for understanding LLM.

\section*{Limitations}
The case studies rely on projecting vectors in vocabulary space, which is widely used in previous studies \cite{elhage2021mathematical,ram2022you,geva2022transformer,dar2022analyzing}. While the results are empirically interpretable, the theories of this method are incomplete. Therefore, we utilize this method in our case studies and supplement our findings with additional methods to strengthen our conclusions, thus enhancing their persuasiveness.

Another limitation lies in the lack of standardization across various studies regarding attribution methods. Different intervention methods (zero intervention, noise intervention, replace intervention, etc.) may get different results. Apart from causal mediation analysis methods and static interpretation methods, gradient-based methods \cite{sundararajan2017axiomatic} and SHAP values \cite{lundberg2017unified} are also widely utilized for attributing important modules. However, these methods often demand substantial computational resources, rendering them unsuitable for our work.  

A potential risk of our work is that attackers can identify the important neurons and edit these neurons to change the output probability distribution. For instance, instead of reducing the gender bias by setting the neurons' parameters to zero, they can amplify the gender bias professions' probabilities by enlarging the identified neurons in Section 6.2. Hence, it is important to distinguish whether a model is edited, and we leave this exploration in future work.

\section{Acknowledgements}
This work is supported by the project JPNP20006 from New Energy and Industrial Technology Development Organization (NEDO). This work is supported by the computational shared facility and the studentship from the Department of Computer Science at the University of Manchester. 

\bibliography{anthology,custom}
\bibliographystyle{acl_natbib}

\appendix

\clearpage
\section{Four Prompts in Arithmetic Dataset}
\begin{table}[htb]
  \centering
  \scalebox{0.95}{\begin{tabular}{ll}
    \toprule
    \textbf{type} & \textbf{prompt} \\
    \midrule
    addition-1 & The sum of n1 and n2 is \\
    addition-2 & Q: What is n1 plus n2? A:\\
    addition-3 & n1 plus n2 is \\
    addition-4 & n1 + n2 = \\
    \midrule
    subtract-1 & The difference between n1 and n2 is \\  
    subtract-2 & Q: What is n1 minus n2? A: \\
    subtract-3 & n1 minus n2 is \\ 
    subtract-4 & n1 - n2 = \\ 
    \midrule
    multiply-1 & The product of n1 and n2 is \\
    multiply-2 & Q: What is n1 times n2? A: \\
    multiply-3 & n1 times n2 is \\
    multiply-4 & n1 * n2 = \\
    \midrule
    division-1 & The ratio of n1 and n2 is \\
    division-2 & Q: What is n1 divides n2? A:\\
    division-3 & n1 divides n2 is \\
    division-4 & n1 / n2 = \\
    \bottomrule
  \end{tabular}}
  \caption{Four prompts for 2-digit arithmetic operations.}
\vspace{-10pt}
\end{table}

\section{Results of Interventions on Head $14^{19}$}
We conduct the same experiments as discussed in Section 4.2-4.4. The results of head $14^{19}$ is shown in Table 15 (corresponding to Table 5), Table 16 (corresponding to Table 6), and Table 17 (corresponding to Table 8).
\begin{table}[htb]
  \centering
  \scalebox{0.95}{\begin{tabular}{llllll}
    \toprule
        & top99 & top50 & top30 & top20 & top10\\
    \midrule
    mask  & 84.6 & 82.1 & 74.4 & 66.7 & 51.3 \\
    keep & 48.7 & 51.3 & 53.9 & 53.9 & 64.2 \\
    coef & 50\% & 61\% & 67\% & 70\% & 73\% \\
    \bottomrule
  \end{tabular}}
  \caption{Decrease (\%) of accuracy and coefficient score on all 1D/ cases when masking and keeping the top FFN neurons.}
\vspace{-10pt}
\end{table}

In Table 15, when head $14^{19}$ is intervened, coefficient scores of important neurons in deep FFN layers are reduced, causing the accuracy decrease. Also, the top identified neurons contain much information. Interventions on top99 neurons result in an accuracy decrease of 84.6\%. 

\begin{table}[htb]
  \centering
  \scalebox{0.95}{\begin{tabular}{llllll}
    \toprule
        & top99 & top50 & top30 & top20 & top10\\
    \midrule
    coef & 1.3 & 0.9 & 3.2 & 4.9 & 7.0 \\
    \bottomrule
  \end{tabular}}
  \caption{Decrease (\%) of coefficient score when intervening the lowest neuron among important FFN neurons.}
\vspace{-10pt}
\end{table}

The results of Table 16 also demonstrate that among the identified important neurons, the lower neurons can enhance higher neurons' coefficient scores among deep FFN neurons. Therefore, the prediction enhancing stage also exists.

\begin{table}[htb]
  \centering
  \scalebox{0.95}{\begin{tabular}{lllll}
    \toprule
        & M=0 & M=1 & M=2 & M=3\\
    \midrule
    number & 51,980 & 10,426 & 1,953 & 510 \\
    acc  & 97.5 & 82.1 & 30.8 & 25.7 \\
    \bottomrule
  \end{tabular}}
  \caption{Decrease (\%) of accuracy on 1D/ cases when intervening hidden-interpretable neurons.}
\vspace{-10pt}
\end{table}

In Table 17, The hidden-interpretable neurons in shallow FFN layers are important for 1D/ cases (e.g. "72/8="). When intervening 10,426 hidden-interpretable shallow FFN neurons, the accuracy reduces 82.1\%. For comparison, we randomly intervene 10,426 FFN neurons in shallow FFN layers, and the interventions only cause a decrease of 5.1\%.

Overall, head $14^{19}$ shares the same mechanism with head $17^{22}$. Head $14^{19}$ stores important parameters for division operations, while head $17^{22}$ is responsible for addition and subtraction.

\section{Results of Interventions in GPT-J}
We conduct the same experiments in Section 3.3 in GPT-J. The accuracy when intervening each head is presented in Table 18.

\begin{table}[htb]
  \centering
  \scalebox{0.95}{\begin{tabular}{lllllll}
    \toprule
        & ori & $7^{0}$ & $13^9$ & $0^{11}$ & $15^{6}$ & $14^{14}$ \\
    \midrule
    all & 74.5 & 63.6 & 64.4 & 65.0 & 68.8 & 70.8 \\
    2D+  & 97.0 & 95.0 & 94.0 & 98.0 & 95.0 & 97.0  \\
    2D-  & 78.6 & 63.6 & \textbf{41.8} & 63.6 & 74.5 & 80.0 \\
    2D*  & 71.0 & \textbf{54.0} & 72.0 & \textbf{53.0} & 59.0 & 72.0 \\
    2D/  & 51.5 & 42.0 & 50.0 & 45.6 & 46.7 & \textbf{34.4}  \\
    \bottomrule
  \end{tabular}}
  \caption{Accuracy (\%) when intervening different heads in GPT-J.}
\vspace{-10pt}
\end{table}

In GPT-J, we also observe that different heads store important parameters for various operations. For instance, the accuracy of 2D- decreases significantly when intervening in head $13^9$, whereas head $14^{14}$ holds significant parameters for 2D/.

Then we apply the CNA method between the original model and the intervened model on head $13^9$ on 2D- cases. The results are shown in Table 19 (corresponding to Table 5), Table 20 (corresponding to Table 6), and Table 21 (corresponding to Table 8).

\begin{table}[htb]
  \centering
  \scalebox{0.95}{\begin{tabular}{llllll}
    \toprule
        & top99 & top50 & top30 & top20 & top10\\
    \midrule
    mask  & 58.4 & 45.9 & 33.4 & 25 & 25 \\
    keep & 37.5 & 50 & 54.1 & 83.4 & 95.9 \\
    coef & 17\% & 20\% & 21\% & 23\% & 29\% \\
    \bottomrule
  \end{tabular}}
  \caption{Decrease (\%) of accuracy and coefficient score when masking and keeping the top FFN neurons.}
\vspace{-10pt}
\end{table}

\begin{table}[htb]
  \centering
  \scalebox{0.95}{\begin{tabular}{llllll}
    \toprule
        & top99 & top50 & top30 & top20 & top10\\
    \midrule
    coef & 1.9 & 1.7 & 1.6 & 1.2 & 1.9 \\
    \bottomrule
  \end{tabular}}
  \caption{Decrease (\%) of coefficient score when intervening the lowest neuron among important neurons.}
\vspace{-10pt}
\end{table}

In Table 19, the top FFN neurons also play a large role in GPT-J. When intervening the top99 neurons, the accuracy decreases 58.4\%. Compared with Llama, the degrees of coefficient decrease and accuracy change are both smaller. In Table 20, when intervening the lowest neuron among the important neurons identified by our CNA method, the deep neurons' coefficient scores decrease.

\begin{table}[htb]
  \centering
  \scalebox{0.95}{\begin{tabular}{lllll}
    \toprule
        & M=0 & M=1 & M=2 & M=3\\
    \midrule
    number & - & 4,272 & 1,228 & 564 \\
    acc  & - & 100.0 & 83.4 & 20.8 \\
    \bottomrule
  \end{tabular}}
  \caption{Decrease (\%) of accuracy when intervening hidden-interpretable neurons.}
\end{table}
\vspace{-10pt}

Results in Table 21 indicate that the hidden-interpretable shallow FFN neurons also exist in GPT-J. When intervening 4,272 neurons, the accuracy decreases 100\%.

Overall, we observe similar results in GPT-J. Similar to Llama, GPT-J also exhibits the presence of four stages: feature enhancing, feature transferring, feature predicting, and prediction enhancing.

\section{Details for Evaluating Gender Bias}
We design eight prompts to find the most common professions causing the gender bias. The prompts are shown in Table 22, where <gend> is "man" or "woman".

\begin{table}[htb]
  \centering
  {\begin{tabular}{l}
    \toprule
    \textbf{prompt} \\
    \midrule
    A <gend> works as a \\
    A <gend> is employed as a \\
    A <gend> holds a job as a \\
    A <gend>'s occupation is \\
    The job of a <gend> is \\
    The work of a <gend> is \\
    The profession of a <gend> is \\
    The work of a <gend> involves \\
    \bottomrule
  \end{tabular}}
  \caption{Eight prompts for gender bias professions.}
\vspace{-10pt}
\end{table}

We compute the top100 predictions of each prompt for different genders, and compare the different professions, which are shown in Table 23. These professions contain much gender bias. We then apply our CNA method between cases with different genders under the first prompt, and identify the top18 important neurons causing the difference. Finally, we edit the top18 neurons by setting their parameters to zero, and then compute the perplexity difference between different genders for all prompts in both the original and edited model (results are shown in Table 13).
\begin{table}[htb]
  \centering
  {\scalebox{0.95}{\begin{tabular}{lp{6cm}}
    \toprule
    gend & profession \\
    \midrule
    woman & cleaner, nurse, secretary, domestic helper, maid, reception, seller, server, librarian, pharmacist, translator, beautician, dental assistant, hairdresser, volunteer, bookkeeper \\
    man & police, guard, delivery, labour, driver, machinist, roofer, machine operator, lumberjack, technician, miner, nightwatch, painter, photographer, builder, porter \\
    \bottomrule
  \end{tabular}}}
  \caption{Professions with gender bias.}
\vspace{-10pt}
\end{table} 

\section{Important Heads for 1-Digit, 2-Digit and 3-Digit Operations}
We report the top5 important heads for 1D, 2D and 3D operations in this section. For each operation, the experiments are conducted on the last prompt in Table 14. The results are shown in Table 24-27.

\begin{table}[htb]
  \centering
  \scalebox{0.95}{\begin{tabular}{lllllll}
    \toprule
        & ori & $17^{22}$ & $15^{23}$ & $6^{20}$ & $13^{15}$ & $14^{19}$ \\
    1D+ & 88.9 & \textbf{47.6} & 82.8 & 84.1 & 84.1 & 84.1 \\
    \midrule
        & ori & $17^{22}$ & $15^{9}$ & $13^2$ & $6^{20}$ & $12^{16}$ \\
    2D+ & 94.5 & \textbf{39.3} & 86.0 & 87.9 & 88.6 & 89.2 \\
    \midrule
        & ori & $17^{22}$ & $8^{10}$ & $15^{23}$ & $12^{16}$ & $15^9$ \\
    3D+ & 96.1 & \textbf{46.4} & 82.7 & 83.5 & 85.2 & 87.2 \\
    \bottomrule
  \end{tabular}}
  \caption{Results of most important heads for 1D+, 2D+, and 3D+. }
\vspace{-10pt}
\end{table}

\begin{table}[htb]
  \centering
  \scalebox{0.95}{\begin{tabular}{lllllll}
    \toprule
        & ori & $17^{22}$ & $16^1$ & $15^{23}$ & $2^{26}$ & $13^2$ \\
    1D- & 82.0 & \textbf{31.0} & \textbf{51.0} & 53.0 & 57.0 & 65.0 \\
    \midrule
        & ori & $16^1$ & $17^{22}$ & $13^2$ & $15^{23}$ & $12^{16}$ \\
    2D- & 80.0 & \textbf{33.9} & \textbf{37.9} & 61.8 & 63.3 & 70.6 \\
    \midrule
        & ori & $16^1$ & $17^{22}$ & $15^{23}$ & $13^2$ & $12^{16}$ \\
    3D- & 57.1 & \textbf{19.6} & \textbf{22.9} & 29.3 & 34.3 & 40.7 \\
    \bottomrule
  \end{tabular}}
  \caption{Results of most important heads for 1D-, 2D-, and 3D-. }
\vspace{-10pt}
\end{table}

\begin{table}[htb]
  \centering
  \scalebox{0.95}{\begin{tabular}{lllllll}
    \toprule
        & ori & $3^5$ & $20^{18}$ & $6^{24}$ & $17^{22}$ & $0^{30}$ \\
    1D* & 93.0 & \textbf{85.4} & \textbf{86.7} & 87.3 & 89.2 & 89.2 \\
    \midrule
        & ori & $15^9$ & $14^{19}$ & $17^{22}$ & $20^{18}$ & $3^5$\\
    2D* & 56.9 & \textbf{49.3} & 50.1 & 50.5 & \textbf{50.5} & 51.6 \\
    \midrule
        & ori & $3^5$ & $15^9$ & $14^{19}$ & $13^{19}$ & $2^{14}$ \\
    3D* & 32.8 & \textbf{25.9} & \textbf{29.7} & 30.3 & 31.1 & 31.1 \\
    \bottomrule
  \end{tabular}}
  \caption{Results of most important heads for 1D*, 2D*, and 3D*. }
\vspace{-10pt}
\end{table}

\begin{table}[htb]
  \centering
  \scalebox{0.95}{\begin{tabular}{lllllll}
    \toprule
        & ori & $14^{19}$ & $15^9$ & $21^{24}$ & $6^{24}$ & $16^{21}$ \\
    1D/ & 78.9 & \textbf{35.6} & 61.1 & 65.6 & 67.8 & 68.9 \\
    \midrule
        & ori & $14^{19}$ & $12^1$ & $3^3$ & $16^{21}$ & $1^{29}$ \\
    2D/ & 48.6 & \textbf{13.7} & 30.2 & 31.4 & 36.9 & 38.8 \\
    \midrule
        & ori & $14^{19}$ & $3^3$ & $12^1$ & $6^{22}$ & $15^9$ \\
    3D/ & 19.0 & \textbf{9.67} & 12.7 & 13.0 & 13.3 & 13.7 \\
    \bottomrule
  \end{tabular}}
  \caption{Results of most important heads for 1D/, 2D/, and 3D/. }
\vspace{-10pt}
\end{table}

\end{document}